%% file: ijcai24.tex
\documentclass{article}
\usepackage{ijcai24}

\usepackage{times}
\usepackage{soul}
\usepackage{url}
\usepackage[hidelinks]{hyperref}
\usepackage[utf8]{inputenc}
\usepackage[small]{caption}
\usepackage{graphicx}
\usepackage{amsmath}
\usepackage{amsthm}
\usepackage{booktabs}
\usepackage{algorithm}
\usepackage{algorithmic}
\usepackage[switch]{lineno}
\usepackage[inline]{enumitem}

\usepackage{amssymb}
\usepackage[super]{nth}
\usepackage{subfig}
\usepackage{color}
\usepackage{xcolor}
\usepackage{tabularx,colortbl}
\usepackage{multirow}
\usepackage{adjustbox}

\newcommand{\ie}{i.e., }
\newcommand{\eg}{e.g., }

\newcommand{\method}{TG-ODE}

\usepackage[normalem]{ulem}

\definecolor{heatcolor}{HTML}{d55e00} 
\definecolor{greencolor}{HTML}{029e73} 
\definecolor{benchcolor}{HTML}{0173b2} 

\title{Temporal Graph ODEs for Irregularly-Sampled Time Series}

\author{
    Alessio Gravina$^{1,}$\thanks{Corresponding author.},
    Daniele Zambon$^2$,
    Davide Bacciu$^1$,
    Cesare Alippi$^{2,3}$
\affiliations
    $^1$University of Pisa, Pisa, Italy\\
    $^2$The Swiss AI Lab IDSIA, Universit\`{a} della Svizzera italiana, Lugano, Switzerland\\
    $^3$Politecnico di Milano, Milan, Italy\\
\emails
    alessio.gravina@phd.unipi.it, daniele.zambon@usi.ch, davide.bacciu@unipi.it, cesare.alippi@usi.ch
}

\begin{document}

\maketitle

\begin{abstract}

Modern graph representation learning works mostly under the assumption of dealing with regularly sampled temporal graph snapshots, which is far from realistic, e.g., social networks and physical systems are characterized by continuous dynamics and sporadic observations. To address this limitation, we introduce the Temporal Graph Ordinary Differential Equation (TG-ODE) framework,  which learns both the temporal and spatial dynamics from graph streams where the intervals between observations are not regularly spaced. We empirically validate the proposed approach on several graph benchmarks, showing that TG-ODE can achieve state-of-the-art performance in irregular graph stream tasks.
\end{abstract}

\input{Sections/introduction.tex}
\input{Sections/problem.tex}
\input{Sections/method.tex}
\input{Sections/related_work.tex}
\input{Sections/experiments.tex}
\input{Sections/conclusions.tex}

\section*{Acknowledgments}
This work has been supported by EU-EIC EMERGE (Grant No. 101070918), by the EU NextGenerationEU programme under the funding schemes PNRR-PE-AI (PE00000013) FAIR - Future Artificial Intelligence Research, and 
by the Swiss National Science Foundation project HORD-GNN (FNS 204061).
\bibliographystyle{named}


\end{document}

%% file: Sections/introduction.tex
\section{Introduction}
Representation learning for graphs has been gaining increasing attention over recent years. Such popularity often builds on the fact that complex phenomena are frequently understood as systems of interacting entities described as a graph. For such a reason, learning on graph-structured data through Deep Graph Networks (DGNs) \cite{BACCIU2020203,GNNsurvey} 
has been adopted for solving problems in a variety of fields, such as biology, social science, and sensor networks \cite{MPNN,bioinformatics,social_network,google_maps,gravina_schizophrenia,10026802}.


Graph-based processing methods turned out to be extremely effective in processing spatio-temporal data too \cite{DCRNN,wu2019graph,a3tgcn,gclstm,zambon2022aztest,ivan2022learning,jiang_graph_2022,hmm_tgl,cini2023scalable}. Such a scenario is a setting where the temporal modeling accounts for functional dependencies -- assimilated, in a broad sense, as spatial relationships -- existing among the interacting entities.
Real-world complex problems described as temporal graphs, e.g., those associated with social interactions, call for novel methods that can move beyond the common assumptions found in most of the methods proposed until now. 
Indeed, such problems require dealing with mutable relational information, irregularly and severely under-sampled data.

Some recent works propose to model input-output data relations as a continuous dynamic described by a learnable ordinary differential equation  (ODE), instead of discrete sequences of layers commonly used in deep learning.
Neural ODE-based approaches have been exploited to model non-temporal data
, including message-passing functions for learning node-level embeddings \cite{GDE,grand,pde-gcn,graphcon,gravina2023adgn}. Notably, relying on ODEs has shown promising for modeling complex temporal patterns from irregularly and sparsely sampled data \cite{NeuralODE,ode-irregular,neuralCDE}.

In this paper, we formulate \textit{Temporal Graph Ordinary Differential Equation} (\method{}), a general continuous-time modeling framework for temporal graphs that encompasses some methods from the literature as its specific instances, and demonstrate that \method{} can be an effective design choice to operate with irregularly and sparsely sampled observations. \method{} is designed through the lens of ODEs for effective learning of irregularly sampled temporal graphs. 
With \method{}, a differential equation is learned directly from data to solve a downstream task and predictions are trajectories obtained by numerical integration of the learned ODE. 


The key contributions of this work can be summarized as follows: 
\begin{enumerate}[label=\textbf{(\roman*)}, itemsep=3pt, parsep=0pt, leftmargin=0em, itemindent=2.5em, topsep=3pt]
\item
we introduce \method{}, a general modeling framework suited for handling irregularly sampled temporal graphs; 
\item
we introduce new benchmarks of synthetic and real-world scenarios for evaluating forecasting models on irregularly sampled temporal graphs; and 
\item
we conduct extensive experiments to demonstrate the benefits of our method and show that \method{} outperforms state-of-the-art DGNs on all benchmarks.
\end{enumerate}

Finally, we stress that, other than the outstanding empirical performance achieved by even simple \method{} instances, the framework allows us to reinterpret many state-of-the-art DGNs as a discretized solution of an ODE, thus facilitating their extension to handle graph streams with irregular sampling.

%% file: Sections/problem.tex
\section{Problem statement}\label{sec:problem}
We consider a dynamical system of interacting entities $u\in\mathcal V$, which we refer to as \emph{nodes}, that is described by 
a Cauchy problem defined on an ODE of the form
\begin{equation}\label{eq:sys-model}
\frac{d\mathbf X(t)}{dt} = F(\mathbf X(t),  \mathbf E(t), \mathbf z(t)),
\end{equation}
with initial condition $\mathbf{X}(0)=\mathbf{X}_0$.
System state $\mathbf{X}(t)=\{\mathbf{x}_u(t) : u\in \mathcal V(t)\}$ is a function of time $t$ and collects the node-level states $\mathbf x_u(t)\in\mathbb{R}^{d_x}$ associated with each node $u\in \mathcal V(t)$. The node set $\mathcal V(t)$ is allowed to vary over time. 
We denote the dynamic set of edges encoding the node relations as $\mathcal{E}(t)\subseteq \mathcal{V}(t)\times\mathcal{V}(t)$.
$\mathbf{E}(t)=\{\mathbf{e}_{uv}(t): (u,v)\in \mathcal E\}$ is a set of edge-level attributes $\mathbf{e}_{vu}(t)\in\mathbb{R}^{d_e}$ defining the type or strength of the interaction between nodes $u,v\in\mathcal{V}(t)$.
The system can also be driven by vector $\mathbf z_u(t)\in\mathbb{R}^c$ accounting for exogenous variables relevant to the problem at hand, such as weather conditions, hour of the day, or day of the week.
Accordingly, for all $u\in\mathcal{V}(t)$, we write
\begin{equation}\label{eq:sys-model-node}
    \frac{d \mathbf{x}_u(t)}{d t} = F\left(\mathbf{x}_u(t), \mathbf{z}_u(t), \{\mathbf{x}_v(t)\}_{v\in\mathcal{N}_u(t)}, \{\mathbf{e}_{vu}(t)\}_{v\in\mathcal{N}_u(t)}\right),
\end{equation}
to emphasize the local dependencies of node state $\mathbf{x}_u(t)$ at a time $t$ from its neighboring nodes $v \in \mathcal N_u(t)=\{v\in\mathcal V(t): (v,u) \in\mathcal E(t)\}$ at the corresponding time.

\begin{figure}
    \centering
    \includegraphics[width=0.35\textwidth]{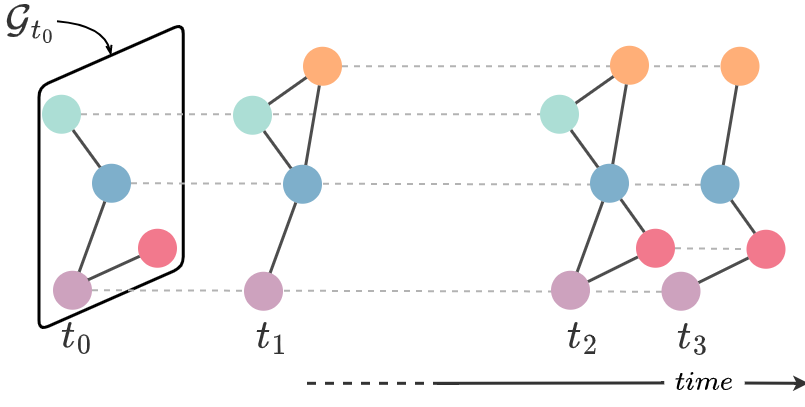}
    \caption{An example of a non-uniform sampling of a temporal graph with snapshots over a set of 5 nodes.}
    \label{fig:dtdg}
\end{figure}

We express any solution of ODE \eqref{eq:sys-model} as the temporal graph 
\begin{equation}
\mathcal G(t) = \left(\mathcal{V}(t), \mathcal{E}(t), \mathbf{X}(t), \mathbf{E}(t)\right)
\end{equation}
defined for $t\ge0$. 
However, we assume to observe system \eqref{eq:sys-model} only as a (discrete) sequence of snapshot graphs 
\begin{equation}\label{eq:graph-seq}
\mathcal G=\{\mathcal G_{t_i}:i=0,1,2,\dots,T\}
\end{equation}
that arrive at irregular timestamps, i.e., the sampling is not uniform and, in general,  $t_i - t_{i-1} \neq t_{i+1} - t_i$.
Each snapshot $\mathcal{G}_t = (\mathcal{V}_t, \mathcal{E}_t, \overline{\mathbf{X}}_t, \mathbf{E}_t)$ corresponds to an observation of the system state at a specific timestamp $t\in\mathbb{R}$. 
Edge set $\mathcal{E}_t$ contains the functional relations between the nodes $u,v\in \mathcal{V}_t$ at time $t$ (thus, $\mathcal{E}_t$ can vary over time), while $\overline{\mathbf{X}}_t=\{\overline{\mathbf{x}}_{t,v}:v\in\mathcal{V}_t\}$ gathers the observed node states, with $\overline{\mathbf{x}}_{t,v}$ the state of node $v$ associated to the observed graph snapshot at time $t$. Edge attributes, if present, are denoted as $\mathrm{e}_{u,v}\in\mathbf{E}_t$.
Figure~\ref{fig:dtdg} visually summarizes this concept.

In this paper, we address the problem of learning a model of the differential equation underlying the observed data, which is subsequently exploited to provide estimates of unobserved system's node states and make forecasts.  
To ease readability, in the following we drop the time variable $t$.

%% file: Sections/method.tex
\section{Temporal Graph Ordinary Differential Equation}
\label{sec:method}

To learn the function $F$ in \eqref{eq:sys-model-node}, we consider a family of models 
\begin{equation}\label{eq:pred-model}
    f_\theta(\mathbf{x}_u,\mathbf z, \{\mathbf{x}_v\}_{v\in\mathcal N_u}, \{\mathbf{e}_{vu}\}_{v\in\mathcal N_u})
\end{equation}
parameterized by vector $\theta$, and optimized so that the solution $\hat{\mathbf x}$ of the differential equation
\begin{equation}\label{eq:pred-ode}
   \frac{d\mathbf{x}_u}{dt} = f_\theta\left(\mathbf{x}_u,\mathbf{z}_u, \{\mathbf{x}_v\}_{v\in\mathcal N_u}, \{\mathbf{e}_{vu}\}_{v\in\mathcal N_u}\right),\quad \forall u\in\mathcal V
\end{equation}
minimizes the discrepancy with the observed sequence of graphs in \eqref{eq:graph-seq}.
We follow the message-passing paradigm \cite{MPNN} and instantiate \eqref{eq:pred-ode} as 
\begin{equation}\label{eq:tg-ode}
    \frac{d \mathbf{x}_u}{d t} = 
    \phi_U\left(\mathbf{x}_u,  \mathbf{z}_u, \rho\left(\{\phi_M(\mathbf{x}_u,\mathbf{x}_v, \mathbf{e}_{vu})\}_{v\in\mathcal{N}_u}\right)\right),
\end{equation}
with message function $\phi_M$, aggregation operator $\rho$ (e.g., the mean), and update function $\phi_U$. Note that functions $\phi_U, \phi_M,$ and $\rho$ have learnable parameters $\theta$, which are shared across all nodes and time steps. An example of a message passing operator is when $\phi_M(\mathbf{x}_u,\mathbf{x}_v, \mathbf{e}_{vu})=\mathbf{x}_v$, $\rho$ is the mean, and $\phi_U$ is a dense feed-forward network of its inputs.
We refer to the above framework in \eqref{eq:tg-ode} as \textit{Temporal Graph Ordinary Differential Equation} (\method{}). 

We observe that, since our framework relies on ODEs, it can naturally deal with snapshots that arrive at an arbitrary time. Indeed, the original Cauchy problem can be divided into multiple sub-problems, one per snapshot in the temporal graph.
Here, the $i$-th sub-problem is defined for all $u\in \mathcal V_{t}$ as
\begin{equation}\label{eq:tg-ode-split}
    \begin{cases}
    \frac{d \mathbf{x}_u}{d t} = 
    \phi_U\left(\mathbf{x}_u,  \mathbf{z}_u, \rho\left(\{\phi_M(\mathbf{x}_u,\mathbf{x}_v, \mathbf{e}_{vu})\}_{v \in\mathcal{N}_u}\right)\right),
    \\
    \mathbf x_u(0) = \psi(\overline{\mathbf x}_{t_{i-1},u}, \hat{\mathbf x}_{u}(t_{i-1})) 
    \end{cases}
\end{equation}
in the time span between the two consecutive timestamps, \ie $t \in [t_{i-1}, t_{i}]$, 
where $\psi$ is a function that combines the $i$-th observed state of the node $u$ related to the snapshot graph $\mathcal G_{t_{i-1}}$ in~\eqref{eq:graph-seq} (\ie $\overline{\mathbf x}_{t_{i-1},u}$) and the prediction $\hat{\mathbf x}_u(t_{i-1})$ obtained by solving \eqref{eq:tg-ode-split} at the previous step. When given, we consider the true -- potentially variable -- topology $\mathcal E(t)$ to define the neighborhoods for $t\in[t_{i-1},t_i]$, otherwise, we set $\mathcal E(t)\equiv\mathcal E_{t_{i-1}}$ for every $t$, i.e., equal to the last observed topology associated with $\mathcal G_{t_{i-1}}$.
Accordingly, we optimize $\theta$ in order to minimize the mean of some loss $\mathcal L$,
\begin{equation}
   \frac{1}{T} \sum_{i=1}^T \frac{1}{|\mathcal V_{t_i}|} \sum_{u\in\mathcal V_{t_i}}  \mathcal L(\overline{\mathbf x}_{t_i,u}, \hat{\mathbf x}_u(t_i))
\end{equation}
where prediction $\hat{\mathbf x}_u(t_i)$ at time $t_i$ is obtained by solving \eqref{eq:tg-ode-split}.



\begin{figure*}[ht]
    \centering
    \includegraphics[scale=0.45]{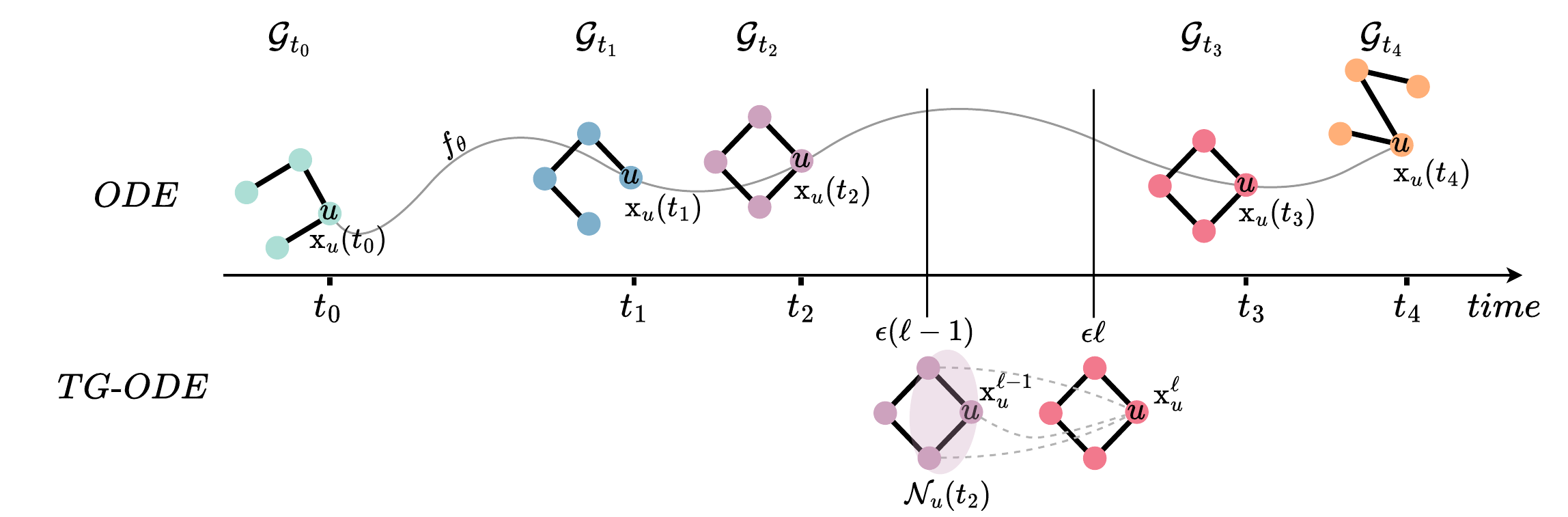} 
    \caption{The continuous processing of node $u$'s state in a discrete-time dynamic graph with irregularly-sampled snapshots over a set of 4 nodes and fixed edge set. At the top, the node-wise ODE function $f_\mathcal{\theta}$ defines the evolution of the states $\mathbf{x}_u(t)$. At the bottom, the discretized solution of the node-wise ODE, which corresponds to our framework \method{}. The node embedding $\mathbf{x}_u^\ell$ is computed iteratively over a discrete set of points by leveraging the temporal neighborhood and self-representation at the previous step.}
    \label{fig:tg-ode}
\end{figure*}

We observe that for most ODEs it is not possible to compute analytical solutions. For such a reason, it is common practice to resort to numerical approximations that leverage discretization strategies (\eg forward Euler's method). In this case, the solution is computed through iterative applications of the method over a discrete set of points in the time interval and, as observed in \cite{NeuralODE,StableArchitecture}, the process can be assimilated to that of a Recurrent Neural Network (RNN). 
For simplicity, here we employ the \textit{forward Euler's method} according to which a solution to~\eqref{eq:tg-ode-split} is obtained by the recursion 
\begin{equation}\label{eq:tg-ode_discretized}
    \mathbf{x}_u^{\ell+1} = \mathbf{x}_u^{\ell} +\epsilon \phi_U\left(\mathbf{x}_u^{\ell},  \mathbf{z}_u(t_{\ell}), \rho\left(\{\phi_M(\mathbf{x}_u^{\ell},\mathbf{x}_v^{\ell}, \mathbf{e}_{vu}^{\ell})\}_{v\in\mathcal{N}_u(t_{\ell})}\right)\right),
\end{equation}
starting from initial condition 
$\mathbf{x}_u^{0}=\mathbf{x}_u(0)=\psi(\overline{\mathbf x}_{t_{i-1},u}, \hat{\mathbf x}_{u}(t_{i-1}))$ and is reiterated until $\epsilon\ell\ge t_i-t_{i-1}$.
In \eqref{eq:tg-ode_discretized}, $\epsilon \ll t_{i}-t_{i-i}$ is the step size, while $\ell$ indicates the generic iteration step, and $t_\ell=t_{i-1} + \epsilon\ell$. 
Finally, a solution $\hat{\mathbf x}(t)$ to~\eqref{eq:tg-ode-split} in the interval $[t_{i-1}, t_i]$ is provided by the discretization $\hat{\mathbf{x}}_u(t_{i-1}+\epsilon\ell)=\mathbf{x}_u^\ell$, for all $\ell$, and interpolated elsewhere. The process is visually summarized at the bottom of Figure~\ref{fig:tg-ode}.

We acknowledge that not all resulting ODEs allow unique solutions and yield numerical stable problems. Generally, numerical stability is associated with the ODE to solve rather than the input data. Thus, a proper design of the considered family of ODEs in \eqref{eq:tg-ode} can prevent stability issues.
Indeed, the solution of a Cauchy problem exists and is unique if the differential equation is uniformly Lipschitz continuous in its input and continuous in $t$, as states in the Picard–Lindel\"{o}f theorem~\cite{picard-theo}. Thus, different implementations of \eqref{eq:tg-ode} should address the continuous behavior of the differential equation. We note that this theorem holds for our model if the underlying neural network has finite weights and uses Lipshitz non-linearities,
\eg the tanh
.

By \eqref{eq:tg-ode_discretized} and the generality of the message passing in \eqref{eq:tg-ode}, we observe that \method{} allows us to cast basically any standard DGN through the lens of an ODE for temporal graphs with irregular timestamps. Secondly, we stress that, even though \method{} is solved here by means of the forward Euler's method, other discretization methods can still be utilized. To conclude, our framework can be implemented using the aggregation function that is most suitable for the given task and the discretization method that best fits the computational resources and problem at hand, such as \cite{GAT,gine,gread,Eliasof_Haber_Treister_2024}. As a demonstration of this, in Section~\ref{sec:experiments} 
we explore the neighborhood aggregation scheme proposed in \cite{tagcn}. Thus, \eqref{eq:tg-ode} can be reformulated as 
\begin{equation}
    \frac{d \mathbf{x}_u}{d t} 
    = \sigma\left(
        \sum_{k=0}^K \,\,\sum_{v\in\mathcal{N}_u^k \cup \{ u \}} 
           \alpha_{u,v}^{(k)}\,{\mathbf{x}_v \theta_{k}}
    \right),
\end{equation}
where  $\sigma$ is an activation function, $K$ is the number of hops in the neighborhood, $\theta_k$ is the $k$-th weight matrix, 
$\mathcal{N}_u^k$ is the $k$-hop neighborhood of $u$, and $\alpha_{u,v}^{(k)}$ is a normalization term. For instance, $\alpha_{u,v}^{(k)}=\left(\hat{d}_v^{(k)}\hat{d}_u^{(k)}\right)^{-1/2}$ weighs according to the degrees $\hat{d}_v^{(k)}$ and $\hat{d}_u^{(k)}$ of  nodes $v$ and $u$ in the $k$-hop graph; other choices can include edge attributes as well.
We note that $\theta_k$ is a parameter specific to the $k$-hop neighborhood of node $u$. Thus, it allows the model to learn different transformation patterns at different distances from the considered node $u$.

%% file: Sections/related_work.tex
\section{Related work}\label{sec:related-work}
\paragraph{Deep Graph Networks for static graphs}
In the static graph domain, most DGNs can be generalized by the concepts introduced by 
MPNN~\cite{MPNN}, which is designed to capture and propagate information between nodes in a graph through a message-passing mechanism. Thus, each node updates its state by exchanging information with its neighboring nodes. 
MPNN can be formulated as 
\begin{equation}
    \label{eq:MPNN}
    \mathbf{x}_u^\ell = \phi_U\left(\mathbf{x}_u^{\ell-1}, \sum_{j\in \mathcal{N}_u} \phi_M\left(\mathbf{x}_u^{\ell-1}, \mathbf{x}_v^{\ell-1}, \mathbf{e}_{uv}\right)\right)
\end{equation}
where $\phi_U$ and $\phi_M$ are respectively the update and message functions, which are responsible for computing neighbor states and updating the node's state accordingly. 
Most of the state-of-the-art DGNs can be derived by \eqref{eq:MPNN}. Indeed, the definition of the message and update function allows implementing DGNs with different properties \cite{GCN,GAT,SAGE,GIN,chebnet,gine,tagcn}.

We can extend the abovementioned DGNs to the domain of temporal graphs by selecting appropriate operators in \eqref{eq:tg-ode} and, in turn, it allows us to tailor the \method{} to exploit relational inductive biases and fulfill given application requirements.
For instance, by considering a graph attentional operator~\cite{GAT} in \eqref{eq:tg-ode}, we can implement an anisotropic message passing within the temporal graph during the update of node states. 

\paragraph{Deep Graph Networks for temporal graphs} 
Given the sequential structure of temporal graphs, a natural choice for many methods has
been to extend Recurrent Neural Networks to graph data. Indeed,
most of the models presented in the literature can be summarized as a combination
of DGNs and RNNs. 

Some approaches adopt a stacked architecture, where DGNs and RNNs are used sequentially \cite{GCRN,evolvegcn}, enabling to separately model spatial and temporal dynamics. Other approaches integrate the DGN inside the RNN \cite{lrgcn,gclstm,GCRN,DCRNN,TGCN,a3tgcn}, allowing to jointly capture the temporal evolution and the spatial dependencies in the graph. We refer to \cite{cini2023graph} and \cite{gravina2023survey} for a deeper discussion.

Differently from these approaches, which are intrinsically designed to deal with regular time series, \method{} can naturally handle arbitrary time gaps between observations. This makes our framework more suitable for realistic scenarios, in which data are irregular over time. 

\paragraph{Continuous Dynamic Models}
NeuralODE (NODE)~\cite{NeuralODE} has emerged as an effective class of neural network models suitable for learning systems' continuous dynamics, drawing a connection between RNNs and ODEs. Despite the similarity with RNNs, such architectures can deal with irregular time series since the continuously-defined dynamics can naturally incorporate data that arrive at arbitrary times \cite{NeuralODE,ode-irregular}. 

Inspired by the NODE approach, GDE~\cite{GDE} links DGNs for static graphs with ODEs. In the static graph domain, ODE-based architectures have been proposed with different aims, such as preserving long-range dependencies~\cite{gravina2023adgn}, reducing the computational complexity of message passing~\cite{dgc,sgc}, and mitigating the over-smoothing phenomena~\cite{pde-gcn,graphcon,kang2024unleashing}.

In the temporal domain, TDE-GNN~\cite{eliasof2024temporal} employs higher-order ODEs to capture the temporal graph dynamic, while NDCN~\cite{ndcn} extends GDE to learn continuous-time dynamics on both static and temporal graphs by diffusing input features until the termination time. MTGODE~\cite{related_first} adopts an ODE-based approach to deduce missing graph topologies from the time-evolving node features in regularly sampled temporal graphs. Differently, \cite{related_fourth} and \cite{related_fifth} propose an ODE-based model in the form of a variational auto-encoder for learning latent dynamics from sampled initial states. To infer missing observations, the methods consider both past and future neighbors' information. This prevents them from being used in an online setting, where data becomes available in a sequential order. 
Lastly, STG-NCDE~\cite{related_second} employs a stacked architecture of two neural controlled differential equations to model temporal and spatial information, respectively. In the STG-NCDE's paper, irregular data are considered, yet they are handled by making them regular via interpolation.

In contrast to these approaches, in this paper we explicitly address irregularly-sampled temporal graphs and we propose a simple model to showcase the effectiveness and efficiency of the TG-ODE framework in working with such data, eliminating the need for additional strategies, such as interpolation. It should be noted that while many of the ODE-based approaches mentioned earlier can be viewed as instances of the introduced TG-ODE framework, our model is specifically designed to demonstrate the benefits of this approach.


%% file: Sections/experiments.tex
\section{Experiments}
\label{sec:experiments}
We provide an empirical assessment of our method against related temporal DGN models from the literature\footnote{We release the code implementing our methodology and reproducing our empirical analysis at \url{https://github.com/gravins/TG-ODE}.}. 
First, we test the efficacy in handling dynamic graphs with irregularly sampled time series by evaluating the models on several heat diffusion scenarios (see 
Section~\ref{sec:heat_diffusion}). Afterward, we assess and discuss the performance on real graph benchmarks on traffic forecasting problems (see 
Section~\ref{sec:real_bench}).  
We report in Table~\ref{tab:general_configs} the grid of hyper-parameters employed in our experiments by each method.
We carried out the experiments on 7 nodes of a cluster with 96 CPUs per node
.

\input{Tables/general_conf}

\subsection{Heat diffusion}
\label{sec:heat_diffusion}

In this section, we focus on simulating the heat diffusion over time on a graph. The data is composed of irregularly sampled graph snapshots 
providing the temperature of the graph's nodes at the given timestamp. We address the task of predicting the nodes' temperature at future (irregular) timestamps.

\subsubsection{Datasets}
In our experiment, we consider a grid graph consisting of 70 nodes, each of which is characterized by an initial temperature $\mathbf x_u(0)$ randomly sampled in the range between $0$ and $0.2$. We randomly alter the initial temperature profile by generating hot and cold spikes located at some nodes. A hot spike is characterized by a temperature between $10$ and $15$, while a cold spike is between $-15$ and $-10$. 
Each altered node has a 40\% chance of being associated with a cold spike and 60\% with a hot spike.
We considered two different experimental scenarios depending on the number of altered nodes. In the first scenario, we alter the temperature of a single node. In the second one, we alter the temperature of one third of the graph's nodes. We will refer to these settings as \textit{single-spike} and \textit{multi-spikes}, respectively.

\begin{figure}
    \centering
    \subfloat[]{\includegraphics[width=0.2\textwidth]{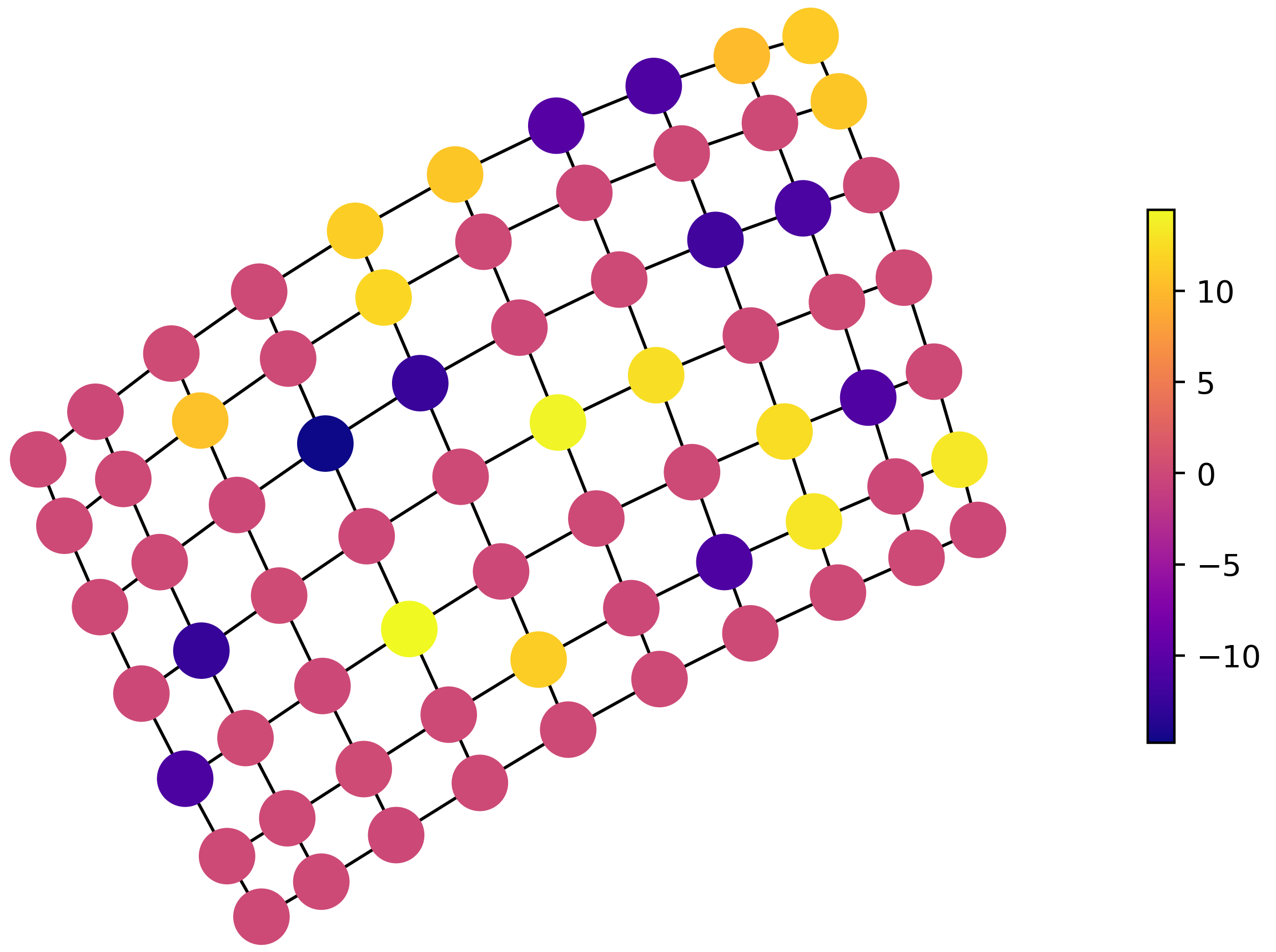}}
    \hspace{1cm}
    \subfloat[]{\includegraphics[width=0.2\textwidth]{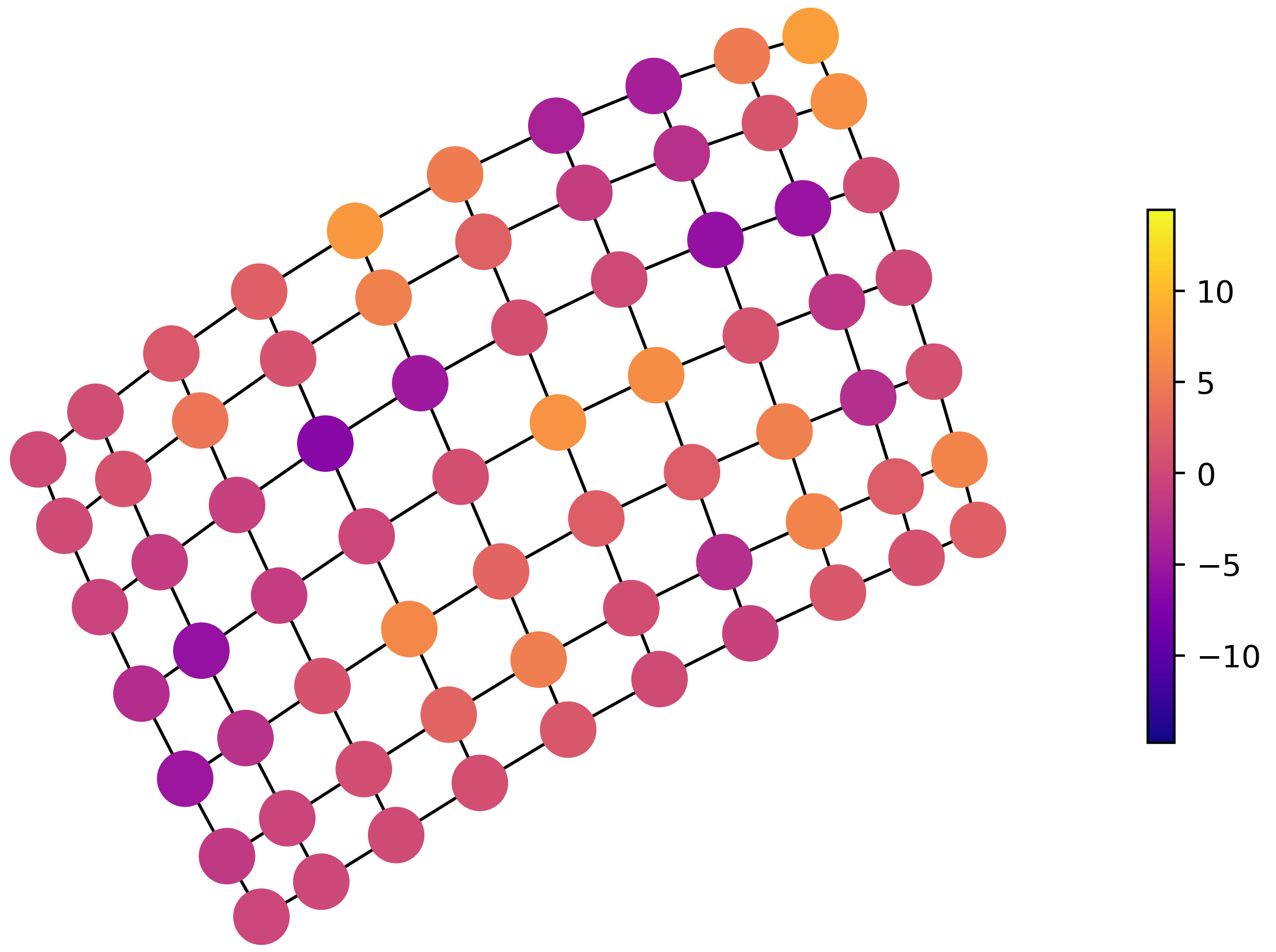}}
    \caption{(a) A grid graph consisting of 70 nodes in which each node is characterized by an initial temperature. Darker colors correspond to colder temperatures, while brighter colors mean warmer temperatures. (b) The heat diffusion simulation is computed through 1000 steps forward Euler's method leveraging $-\mathbf{L}\mathbf{X}(t)$ as diffusion.}
    \label{fig:heat_diffusion}
\end{figure}

We collected the ground truth by simulating the heat diffusion equation through the forward Euler's method with step size $\epsilon=10^{-3}$. Figure~\ref{fig:heat_diffusion} illustrates two snapshot graphs from the simulated heat diffusion.
The training set consists of 100 randomly selected timestamps over the 1000 steps used to simulate the diffusion process. The validation and test sets are generated from two different simulations similar to the one used for building the training set. However, validation and test 
sets are obtained through 500-step simulations, and only 50 of them are kept as validation/test sets. We simulated seven different diffusion functions, \ie $-\mathbf{L}\mathbf{X}(t)$, $-\mathbf{L}^2\mathbf{X}(t)$, $-\mathbf{L}^5\mathbf{X}(t)$, $-\tanh(\mathbf{L})\mathbf{X}(t)$, $-5\mathbf{L}\mathbf{X}(t)$, $-0.05\mathbf{L}\mathbf{X}(t)$, and $-(\mathbf{L}+\mathcal{N}_{0,1})\mathbf{X}(t)$. Here, $\mathcal{N}_{0,1}$ stands for a noise sampled from a standard normal distribution, and $\mathbf{L}$ is the normalized graph Laplacian. 

\subsubsection{\method{} and baseline models}
We explored the performance of \method{} leveraging the aggregation scheme in \cite{tagcn} and the forward Euler's method as discretization procedure, for simplicity. Thus, the nodes' states for the entire snapshot are updated as
\begin{equation}
        \mathbf{X}^\ell = \mathbf{X}^{\ell-1} +\epsilon\sigma\left(\sum_{k=0}^K \mathbf{L}^k \mathbf{X}^{\ell-1} \theta_{k}\right),
\end{equation}
where $K$ corresponds to the number of neighborhood hops and $\theta_{k}$ is the $k$-th weight matrix.
We recall that other choices of aggregation and discretization schemes are possible. We compared our method with six common DGNs for dynamic graphs: 
A3TGCN~\cite{a3tgcn}, DCRNN~\cite{DCRNN}, TGCN~\cite{TGCN}, GCRN-GRU~\cite{GCRN}, GCRN-LSTM~\cite{GCRN}, and NDCN~\cite{ndcn}. We note that whenever we used the NDCN model with embedding dimension set to none (see Table~\ref{tab:general_configs}), the resulting model corresponds to DNND~\cite{ijcai2023p242}.


Moreover, we considered two additional baselines: NODE~\cite{NeuralODE} and LB-baseline. 
NODE represents an instance of our approach that does not take into account node interactions. 
Instead, LB-baseline returns the same node states received as input (\ie,
the prediction of $\overline{\mathbf X}_{t_{i+1}}$ is $\hat{\mathbf{X}}(t_{i+1}) = \overline{\mathbf{X}}_{t_i}$) and
provides a lower bound on the performance we should expect from the learned models. 

We designed each 
model as a combination of three main components. The first is the encoder which maps the node input features into a latent hidden space; the second is the temporal graph convolution (\ie \method{} or the DGN baselines) or the NODE baseline; and the third is a readout that maps the output of the convolution into the output space. The encoder and the readout are Multi-Layer Perceptrons that share the same architecture among all models in the experiments.

To allow all considered baseline models to handle irregular timestamps, we used a similar strategy employed for \method{}. Specifically, we selected the unit of time, $\tau$,  and then we iteratively applied the temporal graph convolution for a number of steps equal to the ratio between the time difference between two consecutive timestamps and the time unit, \ie $\#steps = (t_{i+1} - t_i)/ \tau$. 

We performed hyper-parameter tuning via grid search, optimizing the Mean Absolute Error (MAE). We trained the models using the Adam optimizer for a maximum of 3000 epochs and early stopping with patience of 100 epochs on the validation error. 

\subsubsection{Results}
We present the results on the heat diffusion tasks in Table~\ref{tab:single_spike} and Table~\ref{tab:multi_spike}, using the $log_{10}(\mathrm{MAE})$ as performance metric in both the single-spike and multi-spikes scenarios. The first observation is that \method{} has outstanding performance compared to literature models and the baseline. Despite its simpler architecture, our method produces an error that is significantly lower than the runner-up in each task. In the single-spike setting, \method{} achieves a $log_{10}(\mathrm{MAE})$ that is on average 308\% to 628\% 
better than the competing models in each task.
\input{Tables/heat_uno_spike}
\input{Tables/heat_molti_spike.tex}
Interestingly, not all DGN-based models are capable of improving the results of the LB-baseline. This situation suggests that such approaches attempt to merely learn the mapping function between inputs and outputs rather than learning the actual latent dynamics of the system. Such behavior becomes more evident in the more complex multi-spike scenario. Here, our method achieves up to almost 2080\% 
better $log_{10}(\mathrm{MAE})$ score and more literature models fail in improving the performance with respect to the LB-baseline. These results indicate that capturing the latent dynamics is fundamental, in particular, when the time intervals between observations are not regular over time. We conclude that such methods from the literature might not be suitable for more realistic settings characterized by continuous dynamics and sporadic observations.


Finally, we observe that GCRN-GRU and GCRN-LSTM generate the highest error levels, while DCRNN, NDCN, and NODE are the best among the baselines. Since literature models use RNN architectures to learn temporal patterns, it is reasonable to assume that the poor performance might be due to the limited capacity of RNNs to handle non-uniform time gaps between observations. In contrast, ODE-based models (NODE, NCDN and ours) demonstrate enhanced learning capabilities in this scenario. The performance gap between our model and the considered baselines 
is an 
indication that the diverse spatial patterns learned by different DGN architectures can 
heavily impact the performance of the performed tasks.

\subsection{Graph Benchmarks}\label{sec:real_bench}

This section introduces a set of
graph benchmarks whose objective is to assess traffic forecasting performance from irregular time series; similar to the heat diffusion tasks, we predict the future node values given only the past history.

\subsubsection{Datasets}
We considered six real-world graph benchmarks for traffic forecasting: 
MetrLA~\cite{DCRNN}, Montevideo~\cite{rozemberczki2021pytorch}, PeMS03~\cite{pems03-08}, PeMS04~\cite{pems03-08}, PeMS07~\cite{pems03-08}, and PeMS08~\cite{pems03-08}; we report additional details about the datasets in Table~\ref{tab:graph_benchmark_stats}. We used a modified version of the original datasets where we employed irregularly sampled observations. We will refer to the datasets by using the subscript ``i'' -- e.g., \textbf{MetrLA}$_i$ -- to make apparent the difference from the original versions.

We generated irregular time series by randomly selecting a third of the original graph snapshots for most of the experiments; ratios from 3\% to 94\% are studied in Figure~\ref{fig:ablation}. We considered a temporal data splitting in which 80\% of the previously selected snapshots are used as training set, 10\% as validation set, and the remaining as test set.

\input{Tables/traffic_dataset_stats}

\input{Tables/traffic.tex}

\subsubsection{\method{} and baseline models}
For these experiments, we considered the same models, baseline and architectural choices of the heat diffusion experiments. Since NODE does not take into account interactions between nodes for its predictions, we choose not to include it as a baseline in this scenario. Hyper-parameter tuning has been performed by grid search, optimizing the MAE. Optimizer settings are the same as for the previous experiments. 

\subsubsection{Results}
\begin{figure}[ht]
    \centering
    \includegraphics[scale=0.35, clip, trim=.3cm .6cm .3cm 2.6cm]{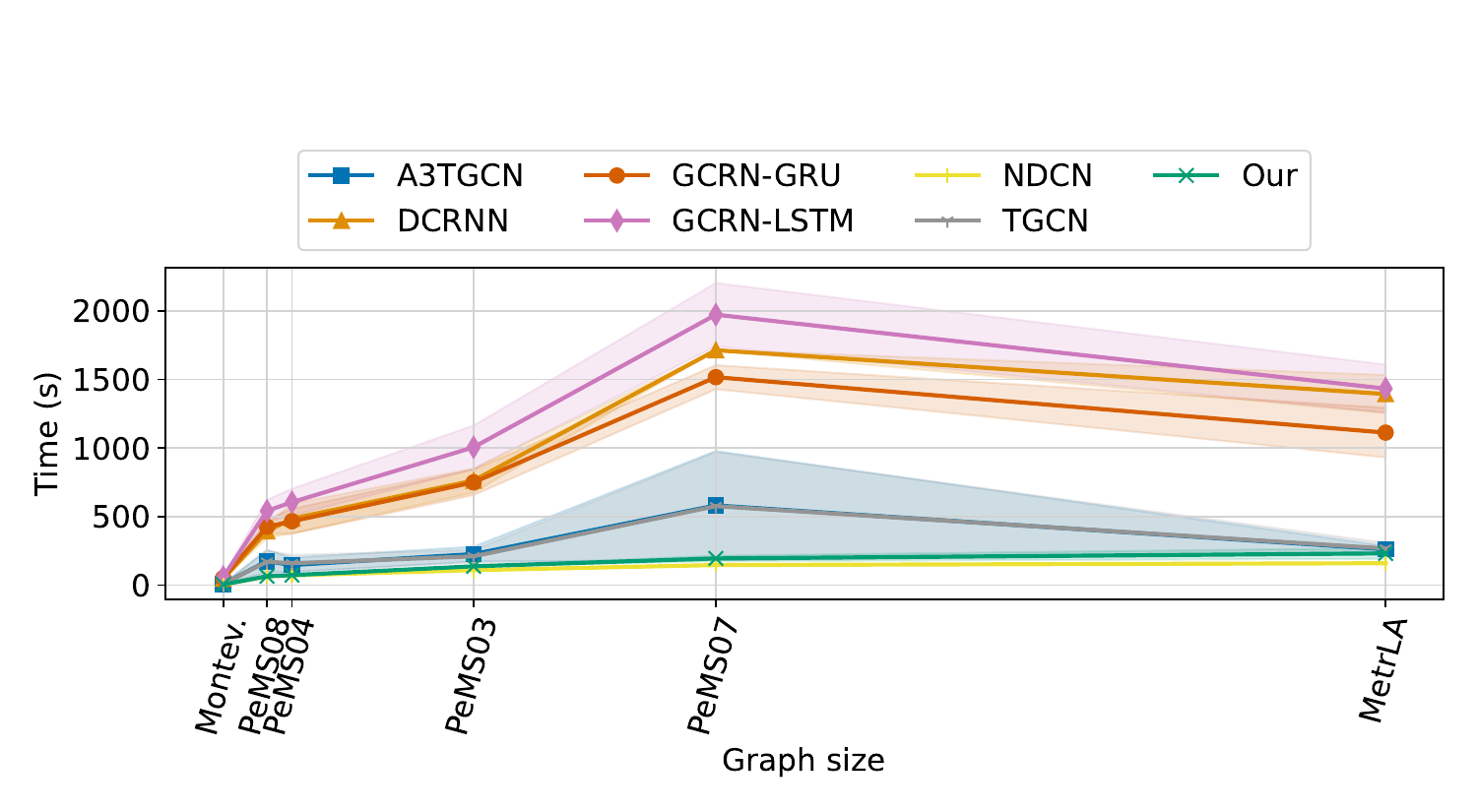}
    \caption{Average time per epoch (measured in seconds) and std computed using an Intel Xeon Gold 6240R CPU @ 2.40GHz. Each time is obtained using 5 neighbor hops (when possible) and embedding dimension equal to 64. The graph size is computed as $size=\#steps * \#edges$.}
    \label{fig:timing}
\end{figure}

Table~\ref{tab:graph_benchmark} reports the traffic forecasting results in
terms of $\mathrm{MAE}$. Similarly to the heat diffusion scenario, \method{} shows a remarkable performance improvement compared to literature models, achieving an MAE that is up to 202\% 
better than the runner-up model. Moreover, as reported in Figure~\ref{fig:timing}
, we observe that \method{} is $2\times$ to $13\times$ faster than the other approaches under test. NDCN is the sole method matching the speed of our approach. However, it's noteworthy that NDCN utilizes only one neighbor hop, thereby simplifying the final computation.

We observe that the baseline performs poorly in these benchmarks, suggesting that such tasks are more complex than the ones based on heat diffusion. 
Despite all DGN models outperforming the LB-baseline, they still produce an error that is on average double 
than that of \method{}, highlighting the added value of our approach when dealing with temporal graphs characterized by irregular sampling.


Finally, we comment that the DCRNN and NDCN suffered from gradient issues in most of the tasks. We believe this is due to 
their inability to learn the latent dynamics of the system when the model
s' outputs are not computed over a regular time series. 

\subsubsection{Impact of the sample sparsity}
To demonstrate the effectiveness of our approach, we study the prediction performance under different sparsity levels. We consider here the PeMS04 dataset. In this analysis, we systematically decreased the number of considered graph snapshots in the time series. This reduction makes the resulting task more challenging than the original one, as the snapshots become more sparse over time -- the expected difference $t_{i+1}-t_{i}$ gets larger. Additionally, the model has fewer data to learn the task, thereby amplifying the task complexity. We generated the irregular time series by randomly selecting 500, 1000, 2000, 4000, 8000, or 16000 graph snapshots from the original dataset (from 3\% to 94\% of the original data), resulting in varying degrees of sparsity. 
For each dataset size, we used an 80/10/10 temporal data split and performed hyper-parameter tuning, as previously done in this section.

\begin{figure}[ht]
    \centering
    \includegraphics[scale=0.4, clip, trim=.4cm .4cm .4cm .4cm]{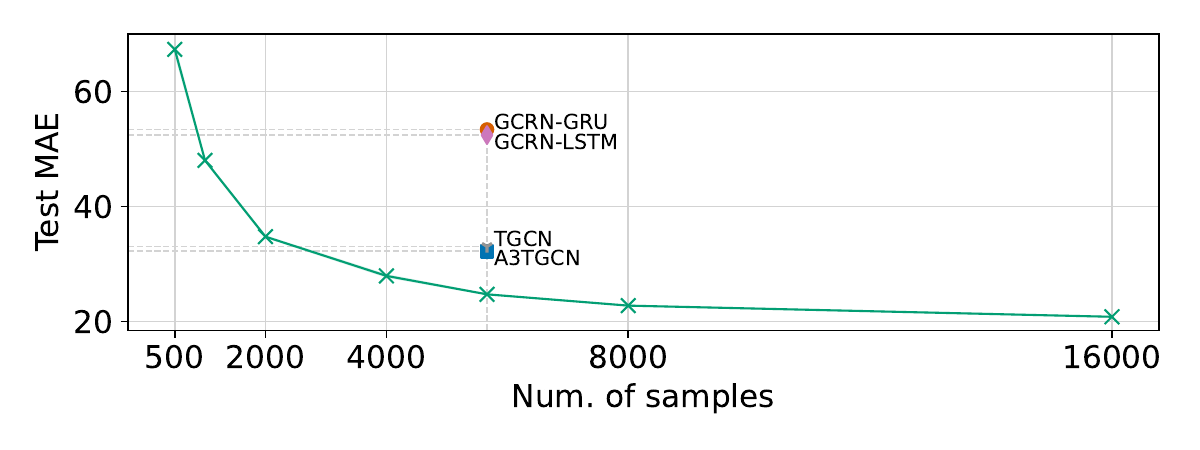}
    \caption{Test $\mathrm{MAE}$ scores and std of \method{} on PeMS04, averaged over 5 runs, for different sparsity levels.}
    \label{fig:ablation}
\end{figure}

Figure~\ref{fig:ablation} illustrates the performance of our method, \method{}, at various degrees of data sparsity. As expected, we observe that as the number of samples increases, the test MAE decreases. 
Notably, \method{} maintains robust performance even with higher degrees of sparsity, with a decrease in performance by only $\sim$7 points when reducing the size from 16000 to 4000 samples. 
While the prediction error is indeed relatively large when considering only 3-7\% of the original data, we comment that it is still substantially better than that of the LB-baseline and comparable to that of the other DGN's which used 33\% of the data. Overall, the model exhibits excellent performance even in situations of high to extreme sparsity (i.e., less than 8000 samples).
The observed outcome supports the effectiveness of \method{}, emphasizing its potential for real-world applications with irregularly sampled temporal graphs. 

%% file: Tables/general_conf.tex
\begin{table}[h]
\caption{The grid of hyper-parameters employed during model selection for the \textcolor{heatcolor}{heat diffusion tasks} (\textcolor{heatcolor}{\emph{Heat}}) and \textcolor{benchcolor}{graph benchmark tasks} (\textcolor{benchcolor}{\emph{Bench}}). 
The $\epsilon$ hyper-parameter is only used by our method (\ie \method), and {\it embedding dim} equal to {\it None} means that no encoder and readout are employed.\label{tab:general_configs}}

\small

\centering
\begin{tabular}{l|c|c}
\hline\toprule
\multirow{2}{*}{\textbf{Hyper-parameters}} & \multicolumn{2}{c}{\textbf{Values}}\\
 &  \textcolor{heatcolor}{\emph{\textbf{Heat}}} & \textcolor{benchcolor}{\emph{\textbf{Bench}}} \\\midrule
Learning rate & \multicolumn{2}{c}{$10^{-2}$, $10^{-3}$, $10^{-4}$}\\
Weight decay &  \multicolumn{2}{c}{$10^{-2}$, $10^{-3}$}\\
$\psi$ & \multicolumn{2}{c}{concat, sum, $\psi(\overline{\mathbf x}, \hat{\mathbf x})=\overline{\mathbf x}$}\\
Activation fun. & \multicolumn{2}{c}{tanh, relu, identity}\\
Embedding dim. & \textcolor{heatcolor}{None, 8} & \textcolor{benchcolor}{64, 32}\\
$\epsilon$ & \textcolor{heatcolor}{$10^{-3}$} & \textcolor{benchcolor}{1, 0.5, $10^{-1}$, $10^{-2}$, $10^{-3}$}\\
\# hops & \textcolor{heatcolor}{5} &  \textcolor{benchcolor}{1, 2, 5} \\\bottomrule\hline
\end{tabular}
\end{table}

%% file: Tables/heat_uno_spike.tex
\begin{table*}[ht]
\centering
\caption{Test $log_{10}(\mathrm{MAE})$ score and std in the single-spike heat diffusion experiments, averaged over 5 separate runs.}\label{tab:single_spike}

\small
\begin{adjustbox}{center}
\begin{tabular}{lccccccc}
\hline\toprule
& $-\mathbf{L}\mathbf{X}(t)$   & $-\mathbf{L}^2\mathbf{X}(t)$ & $-\mathbf{L}^5\mathbf{X}(t)$ & $-\tanh(\mathbf{L})\mathbf{X}$& $-5\mathbf{L}\mathbf{X}(t)$  & $-0.05\mathbf{L}\mathbf{X}(t)$ & $-(\mathbf{L}+ \mathcal{N}_{0,1})\mathbf{X}(t)$ \\\midrule

LB-baseline & -0.557 &      -0.572 &      -0.562 &     -0.538 &       -0.337 &       -0.565 &               -0.837\\ 
NODE
& -2.828$_{\pm0.063}$ & -2.657$_{\pm0.053}$& -2.139$_{\pm0.005}$& -2.711$_{\pm0.136}$& -2.313$_{\pm0.016}$& -3.983$_{\pm0.003}$& -2.059$_{\pm0.005}$ \\\midrule

A3TGCN & -0.834$_{\pm0.145}$ & -0.902$_{\pm0.093}$ & -0.819$_{\pm0.036}$ & -0.890$_{\pm0.035}$ & -1.084$_{\pm0.004}$ & -0.653$_{\pm0.001}$ & -0.781$_{\pm0.094}$\\
DCRNN & -1.320$_{\pm0.163}$ & -0.913$_{\pm0.242}$ & -0.867$_{\pm0.305}$ & -1.273$_{\pm0.075}$ & -1.098$_{\pm0.154}$ & -0.964$_{\pm0.366}$ & -1.150$_{\pm0.375}$\\
GCRN-GRU & -0.474$_{\pm0.232}$ & -0.633$_{\pm0.004}$ & -0.464$_{\pm0.064}$ & -0.621$_{\pm0.047}$ & -0.695$_{\pm0.002}$ & -0.640$_{\pm0.019}$ & -0.490$_{\pm0.094}$\\
GCRN-LSTM & -0.430$_{\pm0.140}$ & -0.323$_{\pm0.019}$ & -0.405$_{\pm0.053}$ & -0.351$_{\pm0.097}$ & -0.511$_{\pm0.157}$ & -0.428$_{\pm0.140}$ & -0.367$_{\pm0.790}$\\

NDCN
& -1.497$_{\pm0.034}$& -1.337$_{\pm0.070}$& -0.350$_{\pm0.328}$& -1.485$_{\pm0.075}$& -1.097$_{\pm0.046}$& -2.408$_{\pm0.183}$& -0.414$_{\pm0.155}$\\

TGCN & -0.825$_{\pm0.108}$ & -0.900$_{\pm0.143}$ & -0.804$_{\pm0.074}$ & -0.834$_{\pm0.149}$ & -1.051$_{\pm0.020}$ & -0.653$_{\pm0.001}$ & -0.781$_{\pm0.094}$ \\\midrule 

Ours & \textbf{-4.087$_{\pm0.171}$} & \textbf{-3.106$_{\pm0.181}$} & \textbf{-2.265$_{\pm0.053}$} & \textbf{-4.166$_{\pm0.140}$} & \textbf{-2.351$_{\pm0.036}$} & \textbf{-4.811$_{\pm0.198}$} & \textbf{-2.069$_{\pm0.001}$}\\


\bottomrule\hline
\end{tabular}
\end{adjustbox}
\end{table*}


%% file: Tables/heat_molti_spike.tex
\begin{table*}[ht]
\centering
\caption{Test $log_{10}(\mathrm{MAE})$ score and std in the multi-spikes heat diffusion experiments, averaged over 5 separate runs.}\label{tab:multi_spike}

\small
\begin{adjustbox}{center}
\begin{tabular}{lccccccc}
\hline\toprule

            & $-\mathbf{L}\mathbf{X}(t)$ & $-\mathbf{L}^2\mathbf{X}(t)$ & $-\mathbf{L}^5\mathbf{X}(t)$ & $-\tanh(\mathbf{L})\mathbf{X}(t)$ & $-5\mathbf{L}\mathbf{X}(t)$ & $-0.05\mathbf{L}\mathbf{X}(t)$ & $-(\mathbf{L}+ \mathcal{N}_{0,1})\mathbf{X}(t)$ \\ \midrule
LB-baseline &  0.490 &       0.517 &       0.552 &      0.523 &        0.256 &        0.561 &                0.666 \\
NODE
& -1.708$_{\pm0.016}$& -1.426$_{\pm0.021}$& -1.093$_{\pm0.004}$& -1.671$_{\pm0.006}$& -1.198$_{\pm0.016}$& -2.749$_{\pm0.016}$& -0.979$_{\pm0.047}$\\
\midrule

A3TGCN & 0.443$_{\pm0.087}$ & 0.244$_{\pm0.124}$ & 0.174$_{\pm0.071}$ & 0.509$_{\pm0.058}$ & 0.187$_{\pm0.010}$ & 0.628$_{\pm0.023}$ & 0.328$_{\pm0.060}$\\
DCRNN & -0.140$_{\pm0.092}$ & -0.143$_{\pm0.111}$ & -0.123$_{\pm0.132}$ & -0.122$_{\pm0.120}$ & -0.421$_{\pm0.227}$ & -0.002$_{\pm0.125}$ & -0.212$_{\pm0.333}$\\
GCRN-GRU & 0.586$_{\pm0.003}$ & 0.614$_{\pm0.004}$ & 0.639$_{\pm0.002}$ & 0.610$_{\pm0.002}$ & 0.440$_{\pm0.003}$ & 0.629$_{\pm0.001}$ & 0.719$_{\pm0.003}$\\
GCRN-LSTM & 0.584$_{\pm0.001}$ & 0.610$_{\pm0.002}$ & 0.637$_{\pm0.002}$ & 0.612$_{\pm0.003}$ & 0.440$_{\pm0.005}$ & 0.631$_{\pm0.002}$ & 0.705$_{\pm0.002}$\\

NDCN
& 0.120$_{\pm0.325}$& -0.070$_{\pm0.056}$& 0.315$_{\pm0.245}$& -0.128$_{\pm0.020}$& 0.146$_{\pm0.107}$& -1.357$_{\pm0.053}$& 0.384$_{\pm0.013}$\\

TGCN & 0.404$_{\pm0.236}$ & 0.313$_{\pm0.072}$ & 0.113$_{\pm0.071}$ & 0.493$_{\pm0.056}$ & 0.113$_{\pm0.086}$ & 0.615$_{\pm0.023}$ & 0.364$_{\pm0.134}$ \\\midrule

Ours & \textbf{-4.259$_{\pm0.037}$} & \textbf{-3.705$_{\pm0.143}$} & \textbf{-1.314$_{\pm0.249}$} & \textbf{-3.572$_{\pm0.010}$} & \textbf{-2.350$_{\pm0.083}$} & \textbf{-4.567$_{\pm0.109}$} & \textbf{-1.021$_{\pm0.002}$}\\

\bottomrule\hline

\end{tabular}
\end{adjustbox}
\end{table*}

%% file: Tables/traffic_dataset_stats.tex
\begin{table}[ht]
\centering
\setlength{\tabcolsep}{4pt}
\caption{Statistics of the original version of the datasets.}\label{tab:graph_benchmark_stats}

\small

\begin{tabular}{lcccc}
\hline\toprule
           & \textbf{\# Steps} & \textbf{\# Nodes} & \textbf{\# Edges} & \textbf{Timespan}\\\midrule
MetrLA     & 34,272 & 207 & 1,515 & 01/03 - 30/06 2012 \\ 
Montevideo &   739  & 675 & 690 & 01/10 - 31/10 2020 \\ 
PeMS03     & 26,208 & 358 & 442 & 01/09 - 30/11 2018 \\ 
PeMS04     & 16,992 & 307 & 209 & 01/01 - 28/02 2018 \\ 
PeMS07     & 28,225 & 883 & 790 & 01/05 - 31/08 2017 \\ 
PeMS08     & 17,856 & 170 & 137 & 01/07 - 31/08 2016 \\ 
\bottomrule\hline
\end{tabular}
\end{table}

%% file: Tables/traffic.tex
\begin{table*}[ht!]
\centering
\caption{Test $\mathrm{MAE}$ score and std in the traffic forecasting setting, averaged over 5 separate runs. $\dagger$ means gradient explosion.}\label{tab:graph_benchmark}

\small

\begin{tabular}{lccccccc}
\hline\toprule

            & \textbf{MetrLA$_i$}         & \textbf{Montevideo$_i$}            & \textbf{PeMS03$_i$}         & \textbf{PeMS04$_i$}          & \textbf{PeMS07$_i$}         & \textbf{PeMS08$_i$} \\ \midrule
LB-baseline    & 58.191                      & 0.442                              & 165.015                     & 211.230                      & 314.710                     & 227.380           \\\midrule  
A3TGCN      & 5.731$_{\pm0.011}$          & 0.378$_{\pm4\cdot10^{-4}}$          & 28.897$_{\pm0.733}$          & 32.221$_{\pm1.355}$           & 38.303$_{\pm0.795}$          & 30.652$_{\pm0.995}$               \\
DCRNN       & $\dagger$                   & 0.332$_{\pm0.001}$                  & 18.652$_{\pm0.136}$          & $\dagger$                    & $\dagger$                     & $\dagger$           \\
GCRN-GRU    & 8.438$_{\pm0.004}$          & 0.332$_{\pm0.001}$                  & 49.360$_{\pm18.619}$         & 53.389$_{\pm4.728}$           & 68.785$_{\pm5.787}$          & 51.787$_{\pm10.872}$               \\
GCRN-LSTM   & 8.440$_{\pm0.009}$          & 0.333$_{\pm0.002}$                  & 62.210$_{\pm0.923}$          & 52.427$_{\pm4.162}$           & 151.824$_{\pm17.654}$        & 80.567$_{\pm24.891}$               \\
NDCN
& 8.471$_{\pm0.022}$ & 0.435$_{\pm0.021}$ & $\dagger$ & 127.202$_{\pm0.334}$ & $\dagger$ & 129.667$_{\pm44.385}$ \\
TGCN        & 5.832$_{\pm0.125}$          & 0.380$_{\pm4\cdot10^{-4}}$          & 28.506$_{\pm0.332}$          & 33.059$_{\pm1.063}$           & 38.750$_{\pm1.429}$          & 33.114$_{\pm1.963}$               \\\midrule
Ours         & \textbf{2.828$_{\pm0.001}$} & \textbf{0.327$_{\pm6\cdot10^{-5}}$} & \textbf{17.423$_{\pm0.012}$} & \textbf{24.739$_{\pm0.014}$}  & \textbf{26.081$_{\pm0.004}$} & \textbf{18.818$_{\pm0.021}$}      \\\bottomrule\hline

\end{tabular}
\end{table*}

%% file: Sections/conclusions.tex
\section{Conclusions}
\label{sec:conclusions}
We have presented \textit{Temporal Graph Ordinary Differential Equation} (\method{}), a new general framework for effectively learning from irregularly sampled temporal graphs. 
Thanks to the connection between ODEs and neural architectures, \method{} can naturally handle arbitrary time gaps between observations, allowing to address a common limitation of DGNs for temporal graphs, \ie the restriction to work solely on regularly sampled data.

To demonstrate the benefits of our approach, we conducted extensive experiments on ad-hoc benchmarks that include several synthetic and real-world scenarios. The results of our experimental analysis show that our method outperforms state-of-the-art models for temporal graphs by a large margin. 
Furthermore, our method benefits from a faster training, 
thus suggesting scalability to large networks. 

Despite being appealing for many realistic application setups, 
we acknowledge that not all resulting ODEs allow unique solutions and yield numerical stability problems, thus requiring some additional care from the user. 

Looking ahead to future developments, we intend to broaden the investigation of more sophisticated numerical methods to solve the learned temporal graph ODE, \eg using adaptive multistep schemes \cite{stableEuler2}. 
Extending the proposed framework to the problem of reconstructing missing data is another interesting research direction to consider.